\documentclass{optica-article}

\journal{opticajournal} 

\articletype{Research Article}

\usepackage{pifont}

\begin{document}

\title{SRMamba: Mamba for Super-Resolution of LiDAR Point Clouds}

\author{Chuang Chen,\authormark{1} Wenyi Ge,\authormark{1,*}}

\address{\authormark{1}College of Computer Science, Chengdu University of Information Technology, Chengdu 610225, China; 

}

\email{\authormark{*}gewenyi15@cuit.edu.cn} 


\begin{abstract*} 
In recent years, range-view-based LiDAR point cloud super-resolution techniques attract significant attention as a low-cost method for generating higher-resolution point cloud data.  However, due to the sparsity and irregular structure of LiDAR point clouds, the point cloud super-resolution problem remains a challenging topic, especially for point cloud upsampling under novel views. In this paper, we propose SRMamba, a novel method for super-resolution of LiDAR point clouds in sparse scenes, addressing the key challenge of recovering the 3D spatial structure of point clouds from novel views. Specifically, we implement  projection technique based on Hough Voting and Hole Compensation strategy to eliminate horizontally linear holes in range image. To improve the establishment of long-distance dependencies and to focus on potential geometric features in vertical 3D space, we employ Visual State Space model and Multi-Directional Scanning mechanism to mitigate the loss of 3D spatial structural information due to the range image. Additionally, an asymmetric U-Net network adapts to the input characteristics of LiDARs with different beam counts, enabling super-resolution reconstruction for multi-beam point clouds. We conduct a series of experiments on multiple challenging public LiDAR datasets (SemanticKITTI and nuScenes), and SRMamba demonstrates significant superiority over other algorithms in both qualitative and quantitative evaluations.

\end{abstract*}

\section{Introduction}
LiDAR plays an indispensable role in environmental sensing systems by accurately capturing the spatial structure of 3D scenes \cite{hu2024rangeldm}, providing reliable 3D environmental information support for autonomous driving \cite{9157799,9577567}, robot navigation and scene reconstruction and localization \cite{10204254,xu2022fast,chen2024vpl}. Due to the insufficient density of low-resolution point clouds, the geometric structure information is significantly missing and degradation, and is difficult to fully characterize the details of the target object and complex topological relationships, failing to achieve the needs of high-precision application scenes. However, high-resolution LiDAR point cloud acquisition devices impose extremely high hardware requirements, and the high cost limits large-scale application and popularization.

\begin{figure}[tbp]
\centering\includegraphics[width=\linewidth]{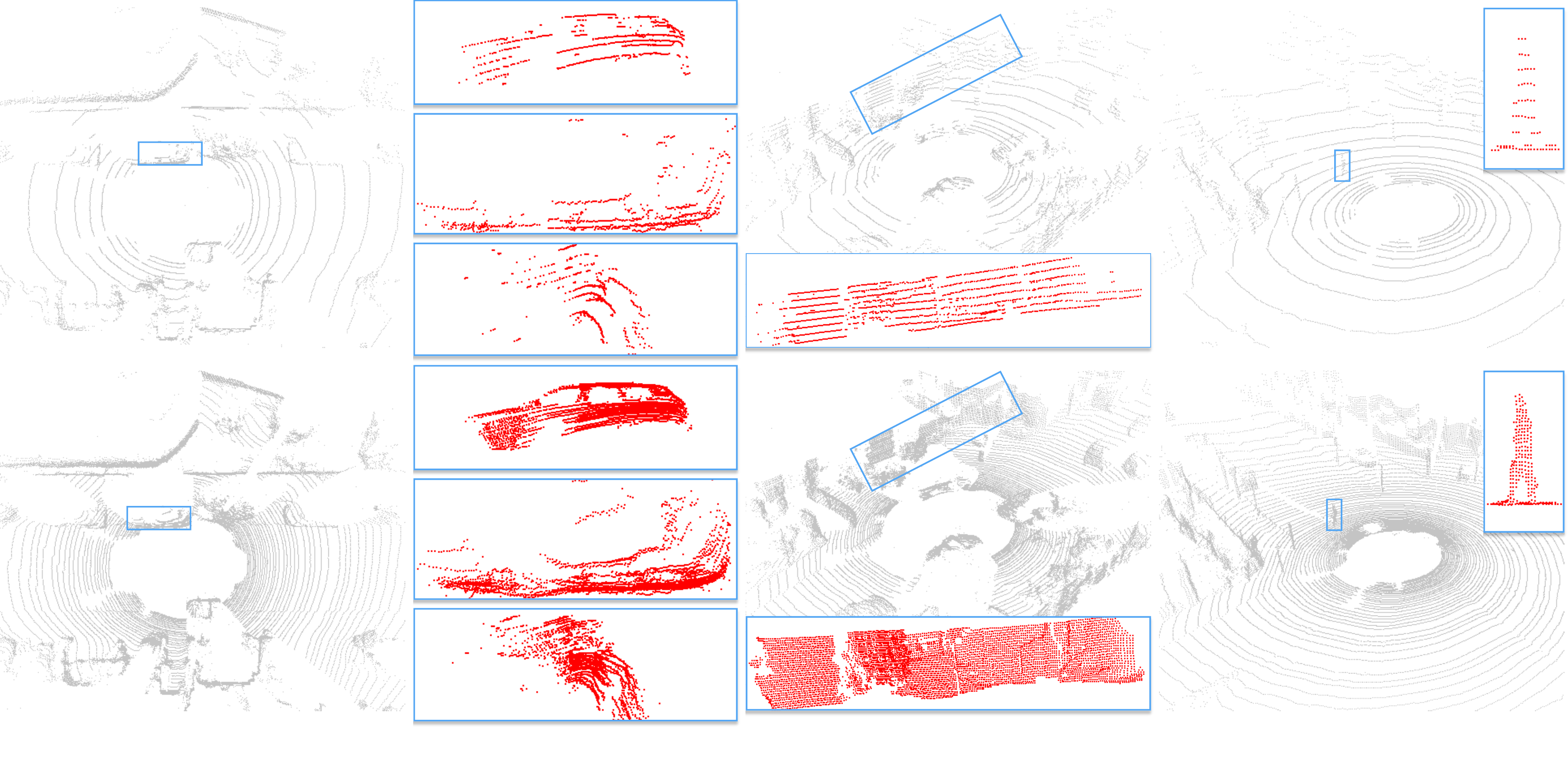}
\caption{Super resolution diagram of the point cloud. The top shows the original 16-line sparse point cloud, with low point density and blurry object outlines; the bottom shows the 64-line point cloud after super-resolution processing, with significantly higher point density, and the structure and details of the object can be clearly reproduced, more accurately reflecting the 3D geometry of the real scene.}
\label{input-output}
\end{figure}

\begin{figure}[htbp]
\centering\includegraphics[width=\linewidth]{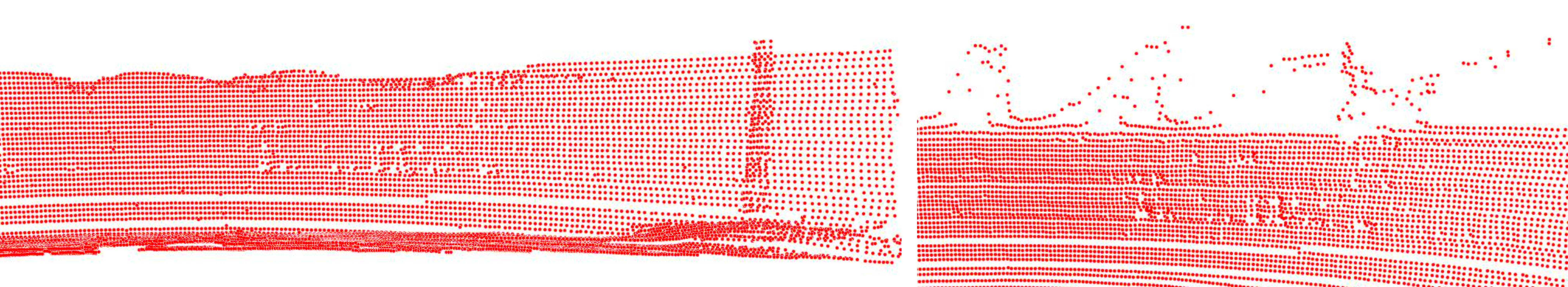}
\caption{Limitations of point cloud super-resolution based on traditional range-view. (1) left: horizontal linear hole. (2) right: offset in the new view.}
\label{limitations-exist}
\end{figure}

To address this challenge, with the rapid development of deep learning techniques, many studies have explores its application in point cloud upsampling \cite{yue20213d,qian2020pugeo}, aiming to improve the resolution and fineness of point cloud data, and to bridge the performance gap at a lower cost, as shown in Fig.~\ref{input-output}. A large number of studies have introduced neural networks to learn the potential spatial features of 3D point clouds and deeply analyze the physical distribution characteristics and geometric structure of LiDAR data \cite{yu2018pu,yifan2019patch,huang2013edge,qian2020pugeo,li2021point}. However, it requires intensive computational resources and is especially unsuitable for super-resolution tasks. Another effective solution is to convert the 3D spatial super-resolution problem into a 2D image super-resolution problem by geometric projection \cite{shan2020simulation,ha2024enhancing,eskandar2022glpu}. Specifically, taking advantage of the deep combination of the physical perceptual properties of range views and the data-driven advantages of neural networks reduces resource consumption, while the attention mechanism performs excellently in capturing details in the field of 2D image super-resolution \cite{niu2020single,zhang2018image,anwar2020deep}. However, 2D features and 3D features possess fundamental differences. Truncation errors during the projection process lead to an irreversible loss of 3D topological structure information,  rendering structural recovery of this region challenging and resulting in the preservation of horizontal linear holes from the range image in the reconstruction process, as shown in Fig.~\ref{limitations-exist}(left). Simultaneously, the attention mechanism cannot model information beyond a finite window and struggles with long-range contextual feature learning \cite{vaswani2017attention}. Consequently, the model overly focuses on structural recovery from the projection viewpoint and struggles to capture spatial structural correlations under new viewpoints, leading to significant coordinate shifts and noise artifacts in the point cloud, as shown in Fig.~\ref{limitations-exist}(right).

Recently, Visual State Space Modeling (VSSM) \cite{liu2024vmamba}, as an efficient computational module, has demonstrated excellent performance in several vision tasks and outperforms Transformer in some scenarios to become one of the cutting-edge technologies in the field of vision \cite{10542538,10822091,10800127}. Its advantages include the ability to efficiently model long-range dependencies with low computational complexity and better inference efficiency. On the other hand, since image patches can be naturally convert to sequence form, VSSM shows a broad application prospect in vision tasks.

In this paper, we propose new network architectures for sparse point cloud super-resolution, motivated by the limitations of range view-based \cite{shan2020simulation} methods and the advantages of VMamba \cite{liu2024vmamba}. To minimize the loss of structure caused by hole pixels, we fill in the blanks using Hough Voting and Hole Compensation mechanism. Meanwhile, using encoding-decoding and skip-connection for multi-scale feature fusion (MSFF), it copes with sparse and scale-inconsistent point cloud inputs. Based on an innovative hybrid RV-VSSM architecture, SRMamba captures local fine-grain features as well as long-distance dependencies in range images, and replaces the quadratic time complexity of the Transformer with linear time complexity. In addition, SRMamba enables recovery of higher resolution 3D point cloud spatial geometries from low-resolution point clouds, maintaining spatial consistency with significantly improving the detail performance of the point cloud, especially in the reconstruction of new viewpoint geometries, showing higher fine-grain. Overall, our contributions are as follows:

\begin{itemize}
\item Propose a point cloud super-resolution network architecture based on VSSM, integrating the multi-scale feature fusion mechanism to effectively improve the ability of the model to perceive sparse input spatial structure, capable of generating high-fidelity high-resolution point cloud scenes with complete structure and rich details.
\item A Hough Voting and a Hole Compensation mechanism are introduced to improve the robustness of the model to hole pixel regions and reduce the position drift and noise interference.
\item Excellent performance on several challenging datasets and high academic and application value compared to existing methods.
\end{itemize}

\section{Related Work}
\subsection{Point Cloud Super-Resolution Based on 3D Space}

Early point cloud up-sampling methods mostly depend on the local geometric features (e.g., normals, density and curvature) of the point cloud for up-sampling, which are highly dependent on the geometric prior \cite{alexa2003computing,lipman2007parameterization,huang2009consolidation}. However, in complex 3D scenes, the irregularity and sparseness of the geometric structure make it difficult to use these methods effectively, especially in LIDAR long-distance detection scenes. Point cloud density diminishes with increasing distance. Concurrently, errors in the geometric prior estimation for edge regions introduce inaccuracies. These combined factors significantly compromise the accuracy and robustness of point cloud reconstruction.

With the development of deep learning, researchers have begun to use neural networks to learn the underlying spatial features of point clouds, gradually moving away from semi-data-driven strategies. PU-Net learns multilevel features of points and implicitly extends the point set and reconstructs it into dense upsampling results \cite{yu2018pu}. Kohei et al. voxelizes the point cloud and introduces sparse convolution to predict high-resolution voxel occupancy \cite{9969624}. Zhang et al. employs a spatial refinement module to predict the offset between the generated coarse dense point cloud and the real one \cite{10688150}. PUGL-Net generates a coarse dense point cloud, further augmented with clustering detail representation \cite{Wang2022Point}. Edge-aware dense convolution (EADC) to reconstruct fine-grained LiDAR scans decouples the up-sampling task into two sub-stages of generation and optimization to fit the object surface \cite{9578328}. Although point cloud processing has achieved positive progress, the inherent sparse and disorderly structure of point clouds, coupled with the lack of explicit structural associations among points, leads to complex neighborhood construction. This complexity, in turn, results in high computational overhead and difficulties in feature modeling, forming the core challenge within this domain. 

\subsection{Point Cloud Super-Resolution Based on Range Images}

Unlike approaches based on prior geometric knowledge and 3D spatial feature modeling, range-view-based point cloud super-resolution techniques achieve a joint optimization of computational efficiency and reconstruction accuracy by deeply integrating well-established 2D vision frameworks with 3D spatial semantics \cite{eskandar2022hals}. The primary objective of image super-resolution is to recover high-frequency details and produce sharper representations from low-resolution (LR) inputs, thereby improving the performance of downstream vision tasks. These methods typically utilize Convolutional Neural Networks (CNNs) to enhance detail fidelity and maintain structural consistency throughout the reconstruction process \cite{9150882,9157565}.

With compactness and high compatibility with LIDAR scanning modes, range images are widely used as an intermediate representation of point cloud super-resolution \cite{meng2019point}. These methods first project the point cloud to image, complete the super-resolution process in the image domain and then back-project to 3D space \cite{10367165}. You et al. performs linear interpolation based on the pixel values of six neighboring points \cite{you2022up}. Tan et al. uses deep convolutional neural networks to improve resolution in image space \cite{shan2020simulation}. Chen et al. \cite{chen2021channel} and TULIP \cite{10657437} performs super-resolution of images through the mechanism of attention. RangeLDM introduces diffusion modeling mechanism on the basis of distance images \cite{hu2024rangeldm}. Despite strong performance metrics in the projection view, the application of these methodologies reveals limitations when extended to a global view. First, an inability to effectively identify hole pixels leads to contamination of robust feature representation. This problem is particularly evident in sparse regions and at object boundaries. Second, range images, being 2D projections of 3D space, result in an overemphasis on local image details while neglecting the inherent 3D spatial structure of the point cloud. Consequently, attempts to generate point clouds from novel viewpoints introduce pseudo-points and cumulative coordinate shifts. Such distortions, including anisotropic stretching, are especially pronounced in long-distance sparse regions. 

In this paper, we focus on generating global high-fidelity high-resolution LiDAR point clouds for large scenes. Considering the frequent interactions between computer vision tasks and sequence modeling, VSSM is rapidly being applied to the image domain \cite{ma2024u,wang2024weak}. Notably, VSSM have higher computational efficiency and larger perceptual range while maintaining sequence modeling capabilities. Unlike previous work, we focus on sequence global dependencies and concentrate more on generating global high-fidelity point clouds rather than regional upsampling.

\begin{figure}[htbp]
\centering\includegraphics[width=\linewidth]{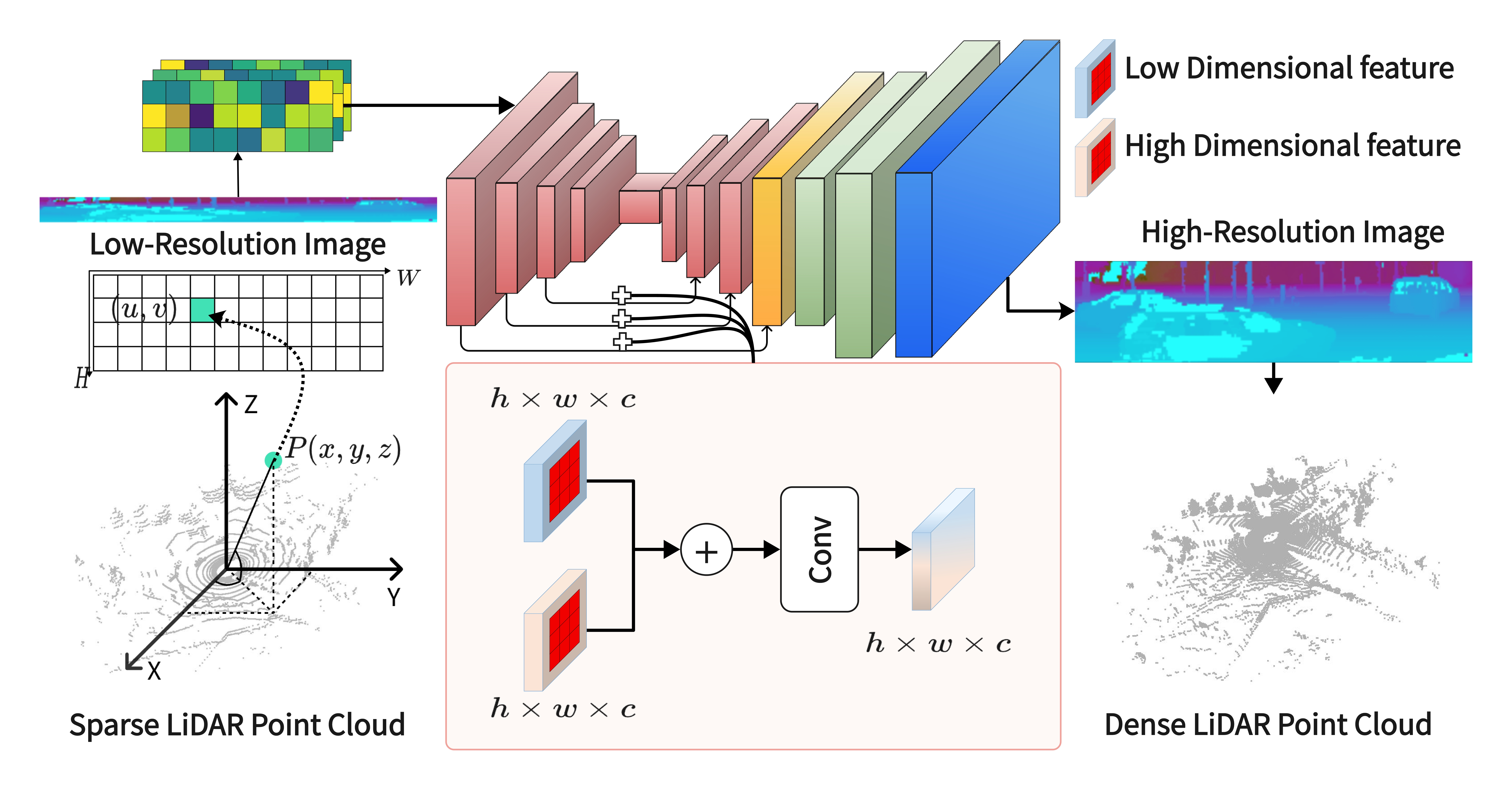}
\caption{Overall framework. The present method takes a sparse point cloud as input, generates a range image, employs a U-Net structure for feature extraction and generates a high-resolution image, back-projects it into the 3D space, and finally generates a high-resolution, high-fidelity representation of the point cloud.}
\label{flowchart}
\end{figure}

\section{Methodology}

We propose a novel LiDAR point cloud super-resolution algorithm, SRMamba, to improve the range view-based point cloud super-resolution algorithm by resolving the noise and structural distortion in the new view region. The algorithm acquires high-quality range images through a hough voting and a hole compensation mechanism, and utilizes the convolution of each anisotropy to compress the image into a compact low-dimensional feature potential space. Multi-scale feature fusion is used to connect high-level features and low-level features to compensate for the loss of high-level semantic information. In the process of training, a bidirectional scanning mechanism is introduced to establish a long distance dependency to obtain a high-resolution 3D point cloud with clear global structure. Fig.~\ref{flowchart} demonstrates the overall process framework.

\subsection{State Space Models}
State Space Models (SSMs) are a mathematical framework used for modeling time series data \cite{gu2023mamba}. The core idea is to use a hidden state vector to describe the dynamic evolution of the system, mapping the input signal $x(t)\in R^{L}$ to an output $y(t)\in R^{L}$. Specifically, a continuous-time SSMs can be represented as a linear ordinary differential equation, as shown in the following equation:

\begin{align}
h^{\prime}(t) =A h(t)+B x(t) \\
y(t)=C h(t)+D x(t)
\end{align}

where the parameters are given by $A\in \mathbb{C}^{N\times N}$,$B,C\in \mathbb{C}^{N}$ for a system with state dimension $N$, alone with a skip connection term $D\in \mathbb{C}$. For seamless integration into neural networks, a time scale parameter $\Delta$ is introduced to discretize the continuous structure using Zero-Order Hold (ZOH)\cite{liu2024vision}:

\begin{align}
    h_{t} &= \bar{A}h_{t-1}+\bar{B}x_{t}
    \\
    y_{t} &= {C}h_{t} + Dx_{t}
\end{align}

where $\bar{A}=e^{\Delta A}$, $\bar{B}=(e^{\Delta A} - I)A^{-1}B$, with $B,C \in \mathbb{R}^{D\times N}$ and $\Delta \in \mathbb{R}^{D}$.

\subsection{Model overview}

\begin{figure}[htbp]
\centering\includegraphics[width=\linewidth]{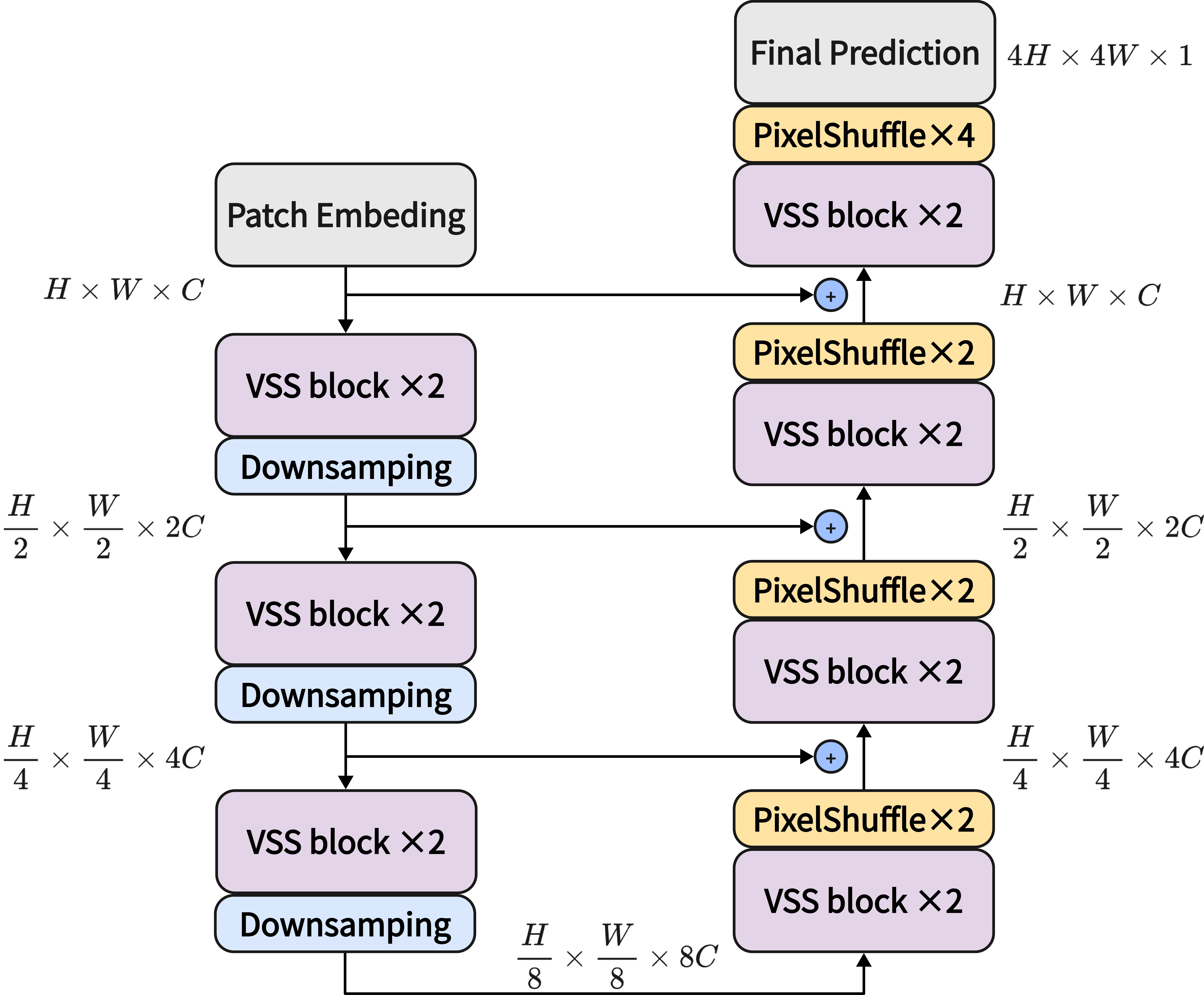}
\caption{SRMamba adopts a hierarchical encoder-decoder architecture, with VSS blocks, downsampling, and PixelShuffle as its core building components. By leveraging multi-scale feature fusion, the model performs super-resolution upsampling tailored to range images.}
\label{network-architecture}
\end{figure}

\subsubsection{Problem Definition}
Given a sparse point cloud $P_{LR}=\{ P_{i}|i=1,2,...,N\}$ acquired by a LiDAR sensor, where each point $P_{i} = (x_{i},y_{i},z_{i})$ represents a 3D spatial coordinate, the objective of the proposed SRMamba framework is to reconstruct a high-quality dense point cloud $P_{HR}$. This process can be formally defined as:

\begin{equation}
    P_{HR} = \mathcal{G}(P_{LR},\theta,scales)
\end{equation}

where $\mathcal{G}$ represents the network structure, and $\theta$ denotes the learnable parameters of the network architecture, $scales$ represents the upsampling factor used to control the resolution of the output point cloud. In this paper, we set it to 4.

\subsubsection{Range View}
Range image is a structured representation of LiDAR point cloud data, with row dimension corresponding to the number of laser beams of LiDAR sensors and column dimension reflecting the distribution of point clouds in the horizontal field of view (FoV) angle \cite{10930660}. However, the original projection method adopts a truncation approach, and the point cloud shows local aggregation, the existence of horizontally linear holes, and the three-dimensional topological relationship is broken. To reduce projection error and optimize image quality, we use a spherical projection method to convert the point cloud into a range image and apply Hough voting to obtain the coordinate offsets of the point cloud on the 2D image, which reduces the projection distortion and geometric error caused by the truncation of the data. Specifically, for each point 
$P_{i} = (x_{i},y_{i},z_{i})$, its spherical coordinates are computed using the following equations:

\begin{equation}
    SC=\left\{\begin{matrix}
      r_{i}=\min(\sqrt{x_{i}^{2}+y_{i}^{2}+(z_{i}-\Delta_{b})^{2}},R_{max})
    \\ 
    {v}_{i}= argmin(|\varphi_{b}-\arctan(\Delta_{b}-z_{i},\sqrt{x_{i}^{2}+y_{i}^{2}})|)
    \\
    
    u_i = \left(1 - (\arctan(y_i, x_i) + \pi)(2\pi)^{-1} \right) W

\end{matrix}\right.
\label{Hough Voting}
\end{equation}

where $\Delta_{b}$ and $\varphi_{b}$ represent the vertical and horizontal correction distances for each laser beam, respectively. They are 1D arrays of length $H$, where $H$ is the number of laser beams. The projected 2D image coordinates are $(v_{i},u_{i})$ , where $v_{i}$ is computed by the $argmin$ operation, which finds the index $v_{i}\in[0,H-1]$ corresponding to the minimum error in the list. To prevent over-correction, we introduce $R_{max}$ as a constraint on the maximum detection range. $W$ represents the pixel width of the range image.

Although Eq.~\ref{Hough Voting} rectifies the horizontally linear holes to a certain extent,
there are still discrete hole regions in the range image due to the sparsity of the input point cloud and this results in broken connections between neighboring regions. In order to address this problem, we propose “Hole Compensation”, which is a technique to diffuse image features to the hole pixels, aiming to fill the empty regions with real features. Specifically, we create a visual window centered at the hole pixels on a dense image optimized for hough voting, and fill the holes with linear average pooling:

\begin{equation}
    \mathcal{} I(u_{i}, v_{i})=\frac{\sum_{(x,y) \in \mathcal{N}(u_{i}, v_{i})} w_{x,y} \cdot I(x,y)}{\sum_{(x,y) \in \mathcal{N}(u_{i}, v_{i})} w_{x,y}}, \quad \text { if } I(u_{i}, v_{i})=\mathrm{NaN}
\end{equation}

Here, $\mathcal{N}(u_{i},v_{i})$ denotes the set of neighboring pixels centered at $(u_{i},v_{i})$, $w_{x,y}$ is the weigh assigned to the neighboring pixel $(x,y)$, and $I(x,y)$ represents the pixel value at $(x,y)$ within the neighborhood.

\subsubsection{Patch Embedding}

Different with the dense three-channel pixel representation of standard RGB images \cite{liu2023uniseg}, there are significant representation differences in range images, which arise from the physical acquisition characteristics of LiDAR-line bundles and FoV angles. To achieve comprehensive recording of point cloud data, employing a larger number of laser beams dictates a need for higher vertical image resolution, while accommodating a wider horizontal FoV necessitates increased horizontal image resolution. To address the anisotropic dimensional distribution (e.g., 16×1024, 64×1024, etc.) and vertical feature sparsity problem specific to range images, the images are mapped into a low-dimensional dense potential space using a feature coding architecture based on an anisotropic convolution kernel (ACK). Specifically, given the input image $I\in R^{C \times H \times W }$, where $C$ denotes the number of channels, and $H$ and $W$ represent the height and width of the image, respectively. we partition the image $I$ into $N$ blocks, each of size $(P_{1},P_{2})$. These blocks are then mapped to a latent representation $E \in R^{D\times (H/scales) \times (W/scales)}$:

\begin{equation}
    E = LayerNorm(Conv2d(I))
    \label{Patch Embeeding}
\end{equation}

\subsubsection{Encoder-Decoder}

\begin{figure}[htbp]
\centering\includegraphics[width=\linewidth]{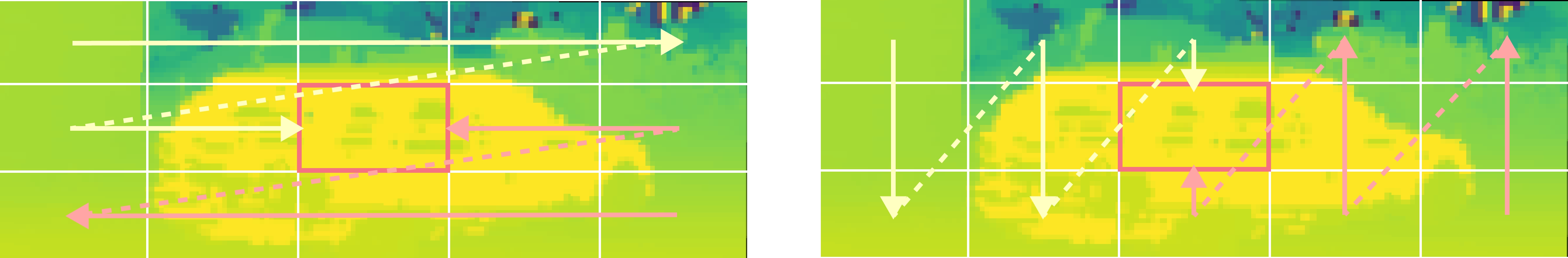}
\caption{A bidirectional scanning mechanism in the spatial domain with scanning directions including left-to-right, right-to-left, top-to-bottom, and bottom-to-top. Each image patch computes the compressed hidden state along the corresponding scan path capturing global context information.}
\label{SS2D}
\end{figure}

SRMamba adopts the asymmetric U-Net network structure, a classical architecture with far-reaching influence in the image processing field \cite{oktay2018attention,wang2022mixed}, as shown in Fig.~\ref{network-architecture}. SRMamba presents the SS2D module, as shown in Fig.~\ref{SS2D}, through two-dimensional multi-directional scanning mechanism, promotes the feature interaction between sparse points at a distance, and realizes the efficient interaction and fusion of global information. SS2D unfolds the input image into sequences along four different paths, processes each sequence in parallel, and finally merges to generate a feature map. Meanwhile, since the range image, as a typical panoramic data, contains rich semantic information of horizontal wide angle in the horizontal direction, the strategy shifts the focus to the vertical dimension of the image.

We employ a 2D backbone network consisting of multiple convolutional modules to efficiently extract multilevel image features. At each stage, we associate the block with multiple stacked VSS modules and apply step-by-step convolution to progressively compress spatial scales and enrich feature representations layer by layer. The VSS module takes a 2D feature map as input, and feeds the result into the core SS2D module to perform 2D multi-directional scanning for efficient global state updates. We then use a linear layer to map the scanned features back to the original feature dimensions and add them to the input features through residual connection \cite{he2016deep}. Afterwards, the output features are again normalized by layers and passed through a feed-forward network (FFN) consisting of a deep convolution (DWConv) and an activation function (SiLU), and finally superimposed with a second residual connection to form the modular output:

\begin{align}
    VSS(X_{in}) &= FFN(LN(SS2D(LN(X_{in}))+X_{in})) + (SS2D(LN(X_{in}))+X_{in})
    \\
    F_{n} &= VSS(F_{i-1}) \quad \text{for } i=1,2,\dots,n
    \\
    VSS_{out} &= F_{n}
    \\
     \mathcal{F}_{encoder}^{l} &= downsamping(VSS_{out})
\end{align}

In the decoding stage, the model adopts a multi-stage progressive up-sampling strategy to recover the resolution of the deepest features step by step, and aligns and fuses them with the shallow features through skip connections and feeds them to the VSS module to make up for the loss of details caused by the resolution reduction. Then, we perform upsampling using the PixelShuffle module, this operation rearranges elements from the channel dimension into the spatial dimension, thereby effectively increasing the image resolution by applying a specified upscale factor $\gamma$. Specially, given an input feature map $(C\times \gamma^{2},h,w)$, PixelShuffle transforms it into $(C, h\times \gamma, w\times \gamma)$, enabling efficient upsampling without relying on interpolation:

\begin{align}
     \mathcal{F}_{decoder}^{l} &= PixelShuffle(VSS(Conv([\mathcal{F}_{decoder}^{l-1}, \mathcal{F}_{encoder}^{l}])))
\end{align}

Finally, 1×1 convolution is using to compress the number of feature map channels to 1, giving an output of a single-channel depth map with a dimension of $(scales\times H) \times (scales\times W)$.

\section{Experiments}

\subsection{Dataset}

To validate the performance of the proposed model, we conduct experiments on two challenging publicly available datasets: KITTI-360 \cite{liao2022kitti} and nuScenes \cite{fong2022panoptic}. The KITTI-360 dataset uses the Velodyne HDL-64E LiDAR to collect 3D structural data of static and dynamic objects in a variety of scenarios, such as cities, villages, and highways \cite{liao2022kitti}. We select 20,000 scans from this dataset as the training set and 2,500 scans as the validation set. The nuScenes dataset, on the other hand, uses the Velodyne HDL-32E LiDAR to acquire 1,000 driving scenarios covering hundreds of thousands of radar scans \cite{fong2022panoptic}. We select 28,130 scans from this dataset as the training set and 6,008 scans as the validation set. And the two datasets are processed with 4 times downsampling to simulate sparse point cloud inputs.

\subsection{Evaluation Metrics}

We construct a multidimensional evaluation system that systematically designs indicators and introduces innovative analysis dimensions to fully demonstrate the comprehensive advantages of the proposed methodology.

Chamfer Distance (CD) \cite{yuan2018pcn} evaluates the point cloud quality in terms of both coverage and completeness dimensions by calculating the mean of the nearest neighbor squared distances between the real and generated point clouds from each other:

\begin{equation}
    CD(S_{\text{pred}}, S_{\text{gt}}) = \frac{1}{N} \sum_{x \in S_{\text{pred}}} \min_{y \in S_{\text{gt}}} \|x - y\|_2^2 
+ \frac{1}{M} \sum_{y \in S_{\text{gt}}} \min_{x \in S_{\text{pred}}} \|y - x\|_2^2
\end{equation}

Intersection over Union (IoU) \cite{kwon2022implicit} computes the geometric similarity of a point cloud by voxelizing the point cloud. We voxelize the point clouds using a voxel size of 0.1 m. $IV$ represents the overlap region between the generated point cloud and the real point cloud in 3D space, and $UV$ represents the total volume covered by the point cloud:

\begin{equation}
    IoU=\frac{IV(S_{pred,S_{gt}})}{UV(S_{pred,S_{gt}})}
\end{equation}

Mean Absolute Error (MAE) \cite{10657437} . In this paper, we generate a point cloud based on the super resolution of the range view, the quality of the range image also determines the quality of the point cloud, and evaluate the similarity between the generated high resolution range image and the real point cloud range image:

\begin{equation}
    MAE=\frac{1}{N}\sum_{i=1}^{N}|x_{i}-\hat {x}_{i}|
\end{equation}

\subsection{Experimental Details}
Range image sizes for KITTI-360 \cite{liao2022kitti} and nuScenes \cite{fong2022panoptic} are 16×1024 and 8×1024, respectively. For optimization, we use AdamW \cite{loshchilov2017decoupled} as the default optimizer with an initial learning rate of 0.005. All models were trained on both datasets for 600 ephemeral sessions using 4× NVIDIA V100 16G GPUs, with batch per GPU sizes of 4 and 8 for each GPU, respectively.

\subsection{Comparison Experiment}

\subsubsection{Qualitative Evaluation}

\begin{figure}[htbp]
\centering\includegraphics[width=\linewidth]{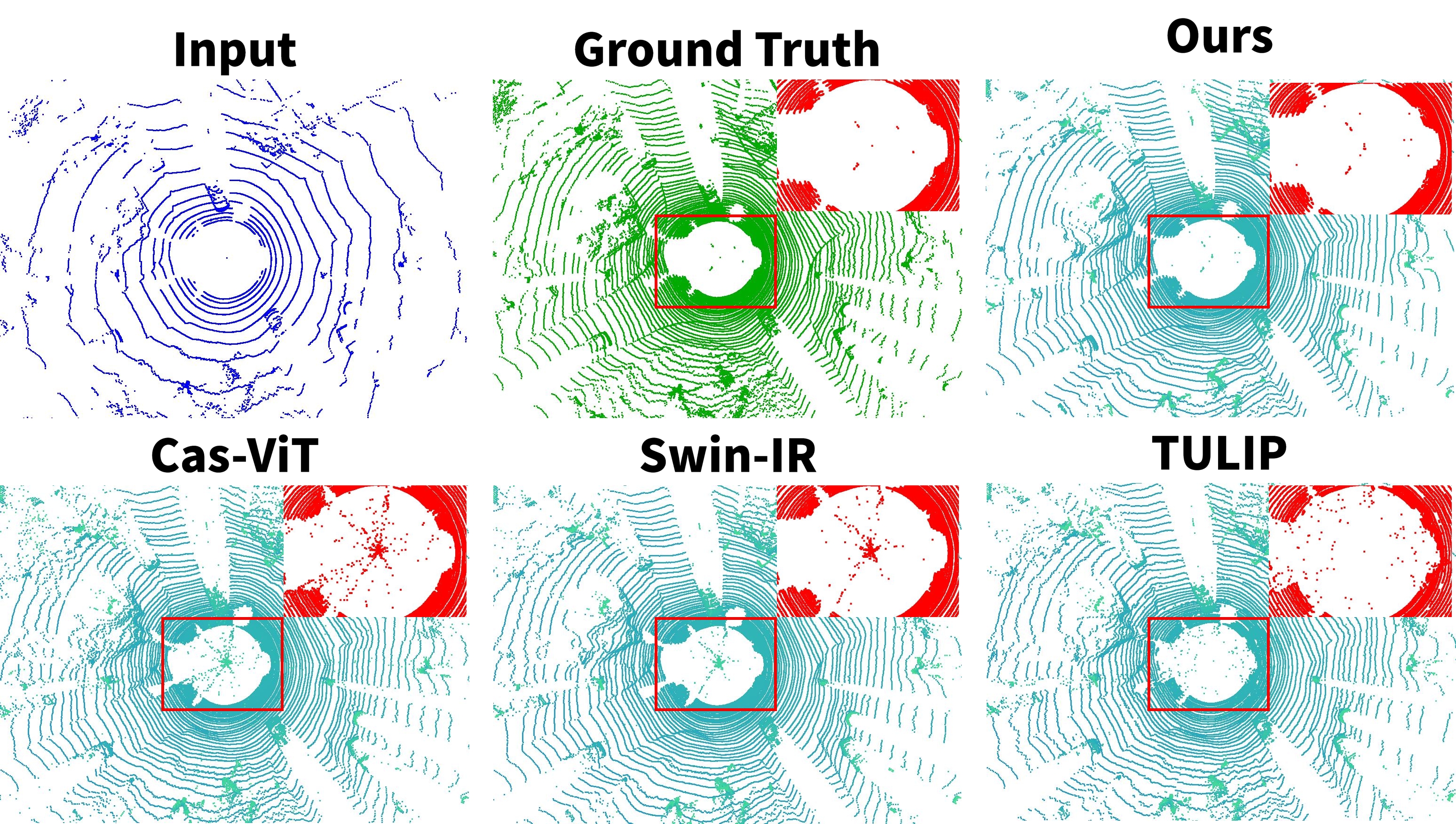}
\caption{Our propose SRMamba method takes sparse point cloud as input to produce realistic high-resolution LiDAR point cloud scenes, effectively recognizing the structural features of the point cloud with smooth structure, clear contours, and rich ensemble details, which is significantly better than other algorithms.}
\label{Global-Comparison}
\end{figure}

Fig.~\ref{Global-Comparison} demonstrates the quality of SRMamba and the competitiveness of the model. We observe the view blindness of Cas-ViT \cite{zhang2024cas}, Swin-IR \cite{liang2021swinir}, and TULIP \cite{10657437} in the center scene of the point cloud, introducing a large amount of noise; in the sparse region, the recovery is inferior, and the reconstructed structures show irregularities and large line fluctuations; while in the complex region, the geometric structures are significantly distorted. In contrast, SRMamba displays results similar to real world ones. Point cloud distribution is uniform, the overall structure is consistent, and there is no extensive point cloud drift or confusion.

\begin{figure}[htbp]
\centering\includegraphics[width=\linewidth]{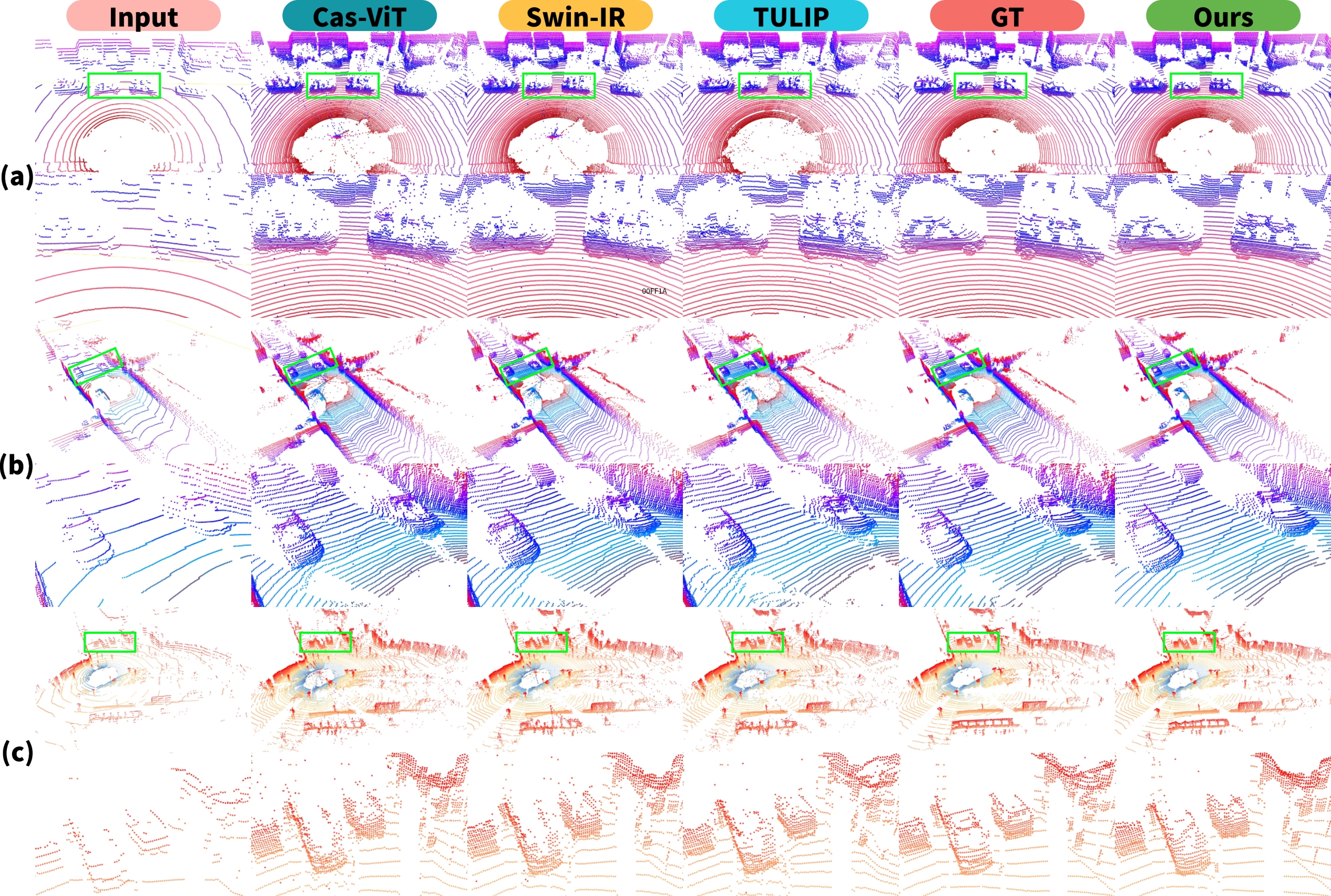}
\caption{Qualitative comparison results of different methods of lidar super-resolution. The zoomed-in details of the area is shown in the green box marked in the above figure in the zoom-in. Comparing with other methods, the 3D point cloud reconstructed by our method is more robust with significantly less noise artifacts.}
\label{Detail-Comparison}
\end{figure}

Fig.~\ref{Detail-Comparison} further demonstrates the performance of reconstruction details in a complex scene, focusing on the ability to recover the vehicle structure. Fig.~\ref{Detail-Comparison}(a) demonstrates the side reconstruction results of multiple cars, with Cas-ViT \cite{zhang2024cas}, Swin-IR \cite{liang2021swinir} and TULIP \cite{10657437} exhibiting significant structural clutter and noise. SRMamba is the only method with clear structure and no significant noise in the occlusion region. In the frontal scene, all methods are able to reconstruct the profile of the car, Cas-ViT \cite{zhang2024cas} and Swin-IR \cite{liang2021swinir} methods are unable to recover the roof structure, and TULIP \cite{10657437} fails to match the LiDAR ground feature lines, as shown in Fig.~\ref{Detail-Comparison}(b). Under the conditions of long distance and highly sparse input point cloud, the reconstruction results of Cas-ViT \cite{zhang2024cas}, Swin-IR \cite{liang2021swinir} and TULIP \cite{10657437} mainly focus on the dense areas on both sides of the truck, ignoring the sparse structure on the top, and the overall contour is incomplete. In contrast, SRMamba can accurately recover the overall shape of the truck, and the reconstruction results are closer to the real scene, as shown in Fig.~\ref{Detail-Comparison}(c).

\begin{figure}[htbp]
\centering\includegraphics[width=\linewidth]{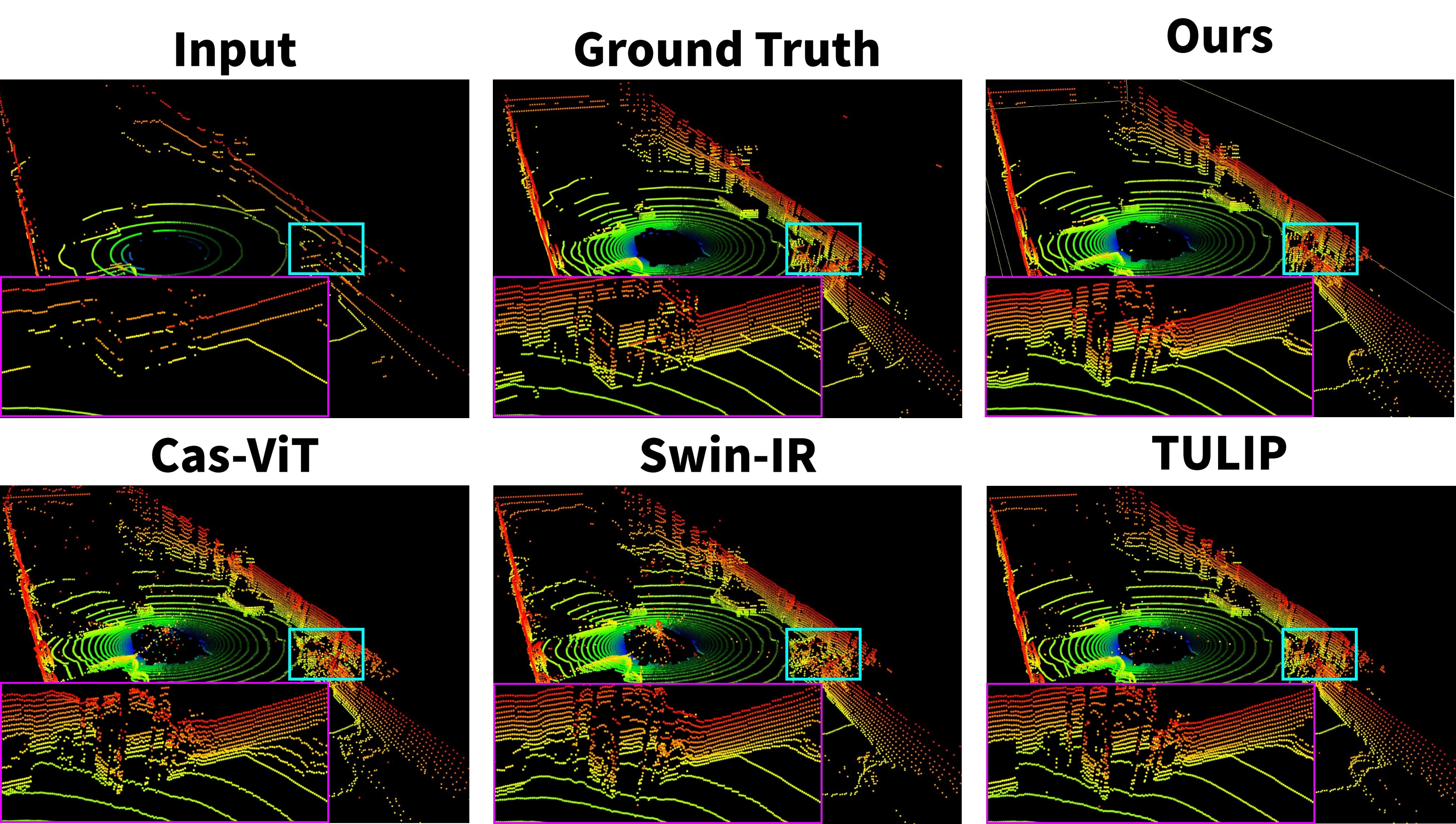}
\caption{Visualization comparison of different methods in sparse to dense point cloud super-resolution task using the nuScenes \cite{fong2022panoptic} dataset, with a black background to highlight sparse geometric details.}
\label{nuScenes}
\end{figure}

Due to the extreme sparsity of the 8-line point cloud and the severe lack of structural information, up-sampling into 32 lines is a highly challenging task. Fig.~\ref{nuScenes} demonstrates that under this sparse condition, Cas-ViT \cite{zhang2024cas}, Swin-IR \cite{liang2021swinir}, and TULIP \cite{10657437} have obvious deficiencies in detail recovery for regions such as walls and building edges in the scene, with problems such as blurry boundaries and collapsing structures. In contrast, SRMamba has clear overall structure and reconstructs continuous wall outlines and a comparably complete edge structure.

\begin{figure}[htbp]
\centering\includegraphics[width=\linewidth]{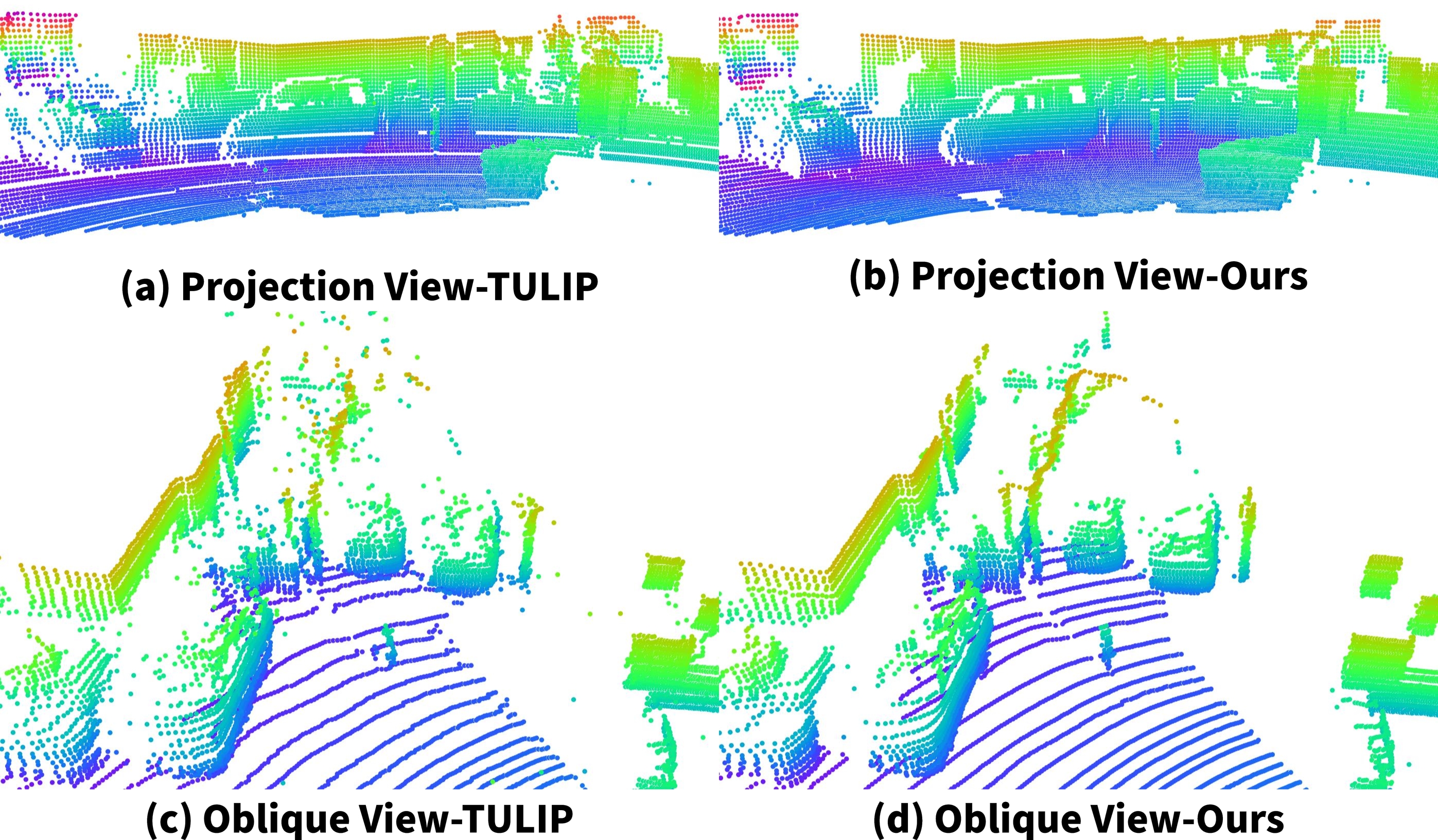}
\caption{Comparison of the spatial structure of the point cloud at the projection view and oblique view. (a) and (c) represent the projection viewpoint and oblique viewpoint of the TULIP \cite{10657437} method. (b) and (d) represent the projection viewpoint and oblique viewpoint of the SRMamba method.}
\label{Oblique-View}
\end{figure}

The range view-based method loses the 3D geometric structure, causing the model to overly focus on regional 2D image features while ignoring the geometric representation of the point cloud in the spatial dimension. It results in a scene with clear geometric structure in the projection view, as shown in Fig.~\ref{Oblique-View} (a), but the point cloud exhibits obvious discretization and broken structure in the new view, as shown in Fig.~\ref{Oblique-View} (c). Meanwhile, the generation of the point cloud scene is consistent with the input range view, and there are horizontal hole regions. In contrast, our proposed SRMamba method, which optimizes the geometric image holes and learns the long-distance dependence through bidirectional scanning mechanism, focuses on the overall structure of the point cloud scene and maintains a clear geometric profile and spatial consistency under multiple viewpoints, as shown in Fig.~\ref{Oblique-View} (b,d).

\subsubsection{Quantitative Evaluation}

Table.~\ref{Comparison of Metrics on the KITTI Dataset} and Table.~\ref{Comparison of Metrics on the nuScenes Dataset} demonstrates the superiority of the proposed SRMamba method over other approaches in both 3D and 2D evaluation metrics. Specifically, SRMamba achieves better performance in terms of all metrics. On the KITTI-360 \cite{liao2022kitti} dataset, SRMamba attains IoU of 0.4548 and CD of 0.0940, outperforming TULIP \cite{10657437} by 9.5\% and 24.3\%, respectively. Similarly, on the nuScenes \cite{fong2022panoptic} dataset, SRMamba continues to lead, maintaining superior accuracy and geometric consistency. 

\begin{table}[htbp]
\centering
\caption{Comparison of Metrics on the KITTI-360 \cite{liao2022kitti} Dataset. SRMamba-T denotes shallower network depth, SRMamba-L indicates deeper network. The best-performing results are highlight in bold.}
\begin{tabular}{cccccc}
\hline
Methods & Dataset & IoU $\uparrow$  & CD $\downarrow$  & MAE$\downarrow$  & Params \\
\hline
Cas-ViT & KITTI-360 & 0.3936 & 0.1483 & 0.0076 & 90.97M \\
Swin-IR & KITTI-360 & 0.4077 & 0.1514 & 0.0078 & 142.58M \\
TULIP & KITTI-360 & 0.4152 & 0.1241 & - & 414.37M \\
\hline
SRMamba-T & KITTI-360 & 0.4389 & 0.1031 & $\mathbf{0.0044}$ & 157.39M \\
\hline
SRMamba-L & KITTI-360 & $\mathbf{0.4548}$ & $\mathbf{0.0940}$ & 0.0048 & 316.10M\\
\hline
\end{tabular}
\label{Comparison of Metrics on the KITTI Dataset}
\end{table}

\begin{table}[htbp]
\centering
\caption{Comparison of Metrics on the nuScenes \cite{fong2022panoptic} Dataset. SRMamba-T denotes shallower network depth, SRMamba-L indicates deeper network. The best-performing results are highlight in bold.}
\begin{tabular}{cccccc}
\hline
Methods & Dataset & IoU $\uparrow$  & CD $\downarrow$  & MAE$\downarrow$  & Params \\
\hline
Cas-ViT & nuScenes & 0.2872 & 1.1624 & 0.0319 & 90.97M \\
Swin-IR & nuScenes & 0.2882 & 1.2527 & 0.0300 & 142.58M \\
TULIP & nuScenes & 0.3048 & 1.0502 & 0.0293 & 414.37M \\
\hline
SRMamba-T & nuScenes & 0.3170 & 1.0196 & 0.0287 & 157.39M \\
\hline
SRMamba-L & nuScenes & $\mathbf{0.3482}$ & $\mathbf{0.9620}$ & $\mathbf{0.0280}$ & 316.10M\\
\hline
\end{tabular}
\label{Comparison of Metrics on the nuScenes Dataset}
\end{table}

\begin{figure}[htbp]
\centering\includegraphics[width=\linewidth]{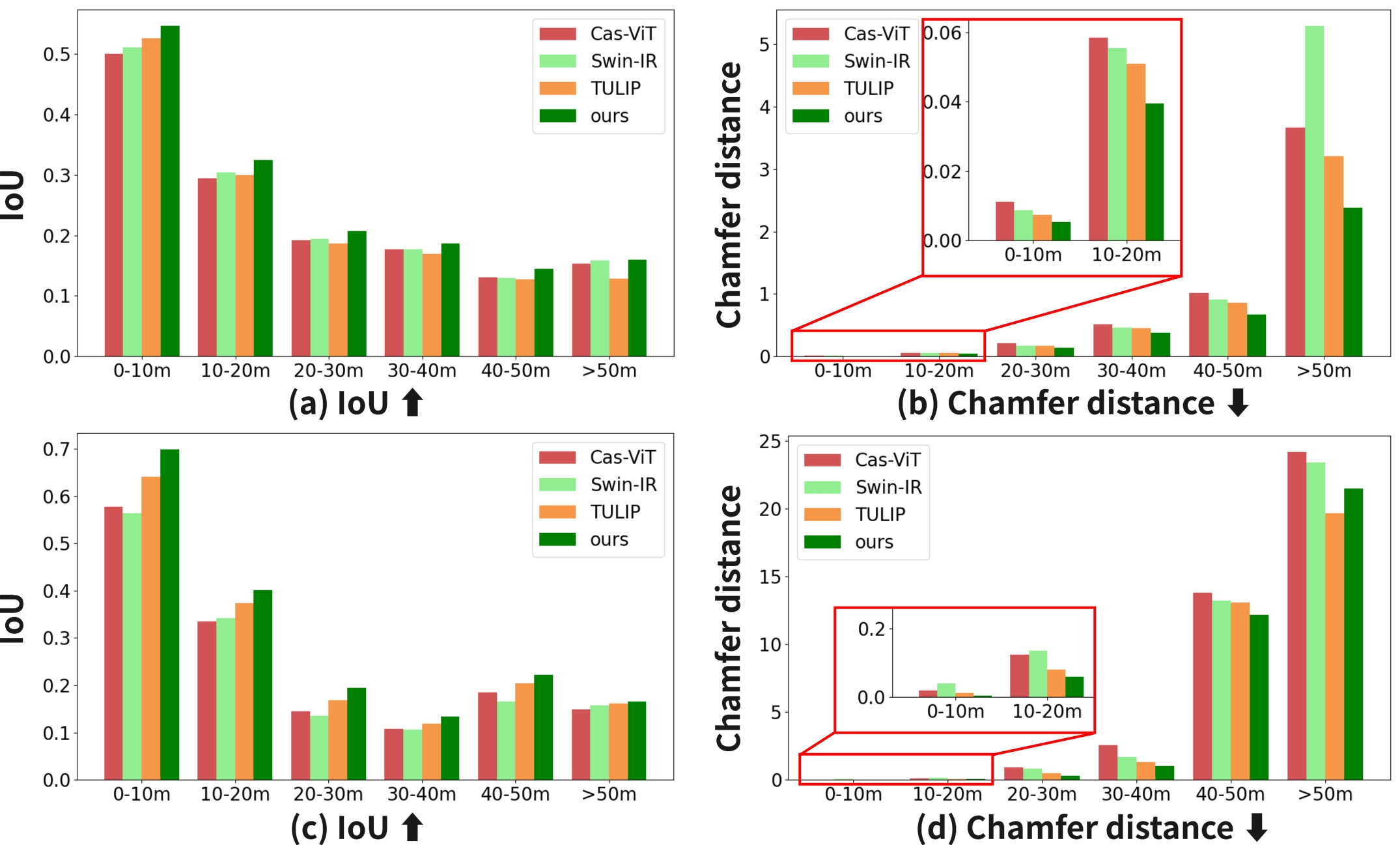}
\caption{(a) and (b) are the 3D metrics metrics visualized by KITTI-360 \cite{liao2022kitti} at different distance segments, respectively. (c) and (d) are the 3D metrics metrics visualized by nuScenes \cite{fong2022panoptic} at different distance segments, respectively.}
\label{range-bar}
\end{figure}

Up-sampling of sparse point clouds is a highly challenging task. With sparser point clouds, up-sampling is more difficult, in addition to the fact the density of the point cloud gradually decreases with increasing distance, which further exacerbates the reconstruction difficulty. In order to achieve a finer evaluation, we analyze the quantitative metrics comparatively in different distance intervals. As shown in Fig.~\ref{range-bar}(a,b), SRMamba exhibits superior performance at all distances, especially in the range of 40-50 meters, which still maintains high accuracy. In the nuScenes \cite{fong2022panoptic} dataset, the distance error between point clouds is significantly higher than that in the KITTI-360 \cite{liao2022kitti} dataset, further highlighting the difficulty of upsampling in sparse scenes, as shown in Fig.~\ref{range-bar}(c,d). Nevertheless, our method still achieves better performance in such complex scenes. 

\subsection{Ablation Study}

\subsubsection{Range Image}

The quality of the range image is critical to the super-resolution of the point cloud based on the range view method. Due to the minor differences on the range image is dramatically amplified after back-projection into 3D space, directly affecting the geometric accuracy and overall structural coherence of the reconstructed point cloud.

\begin{figure}[htbp]
\centering\includegraphics[width=\linewidth]{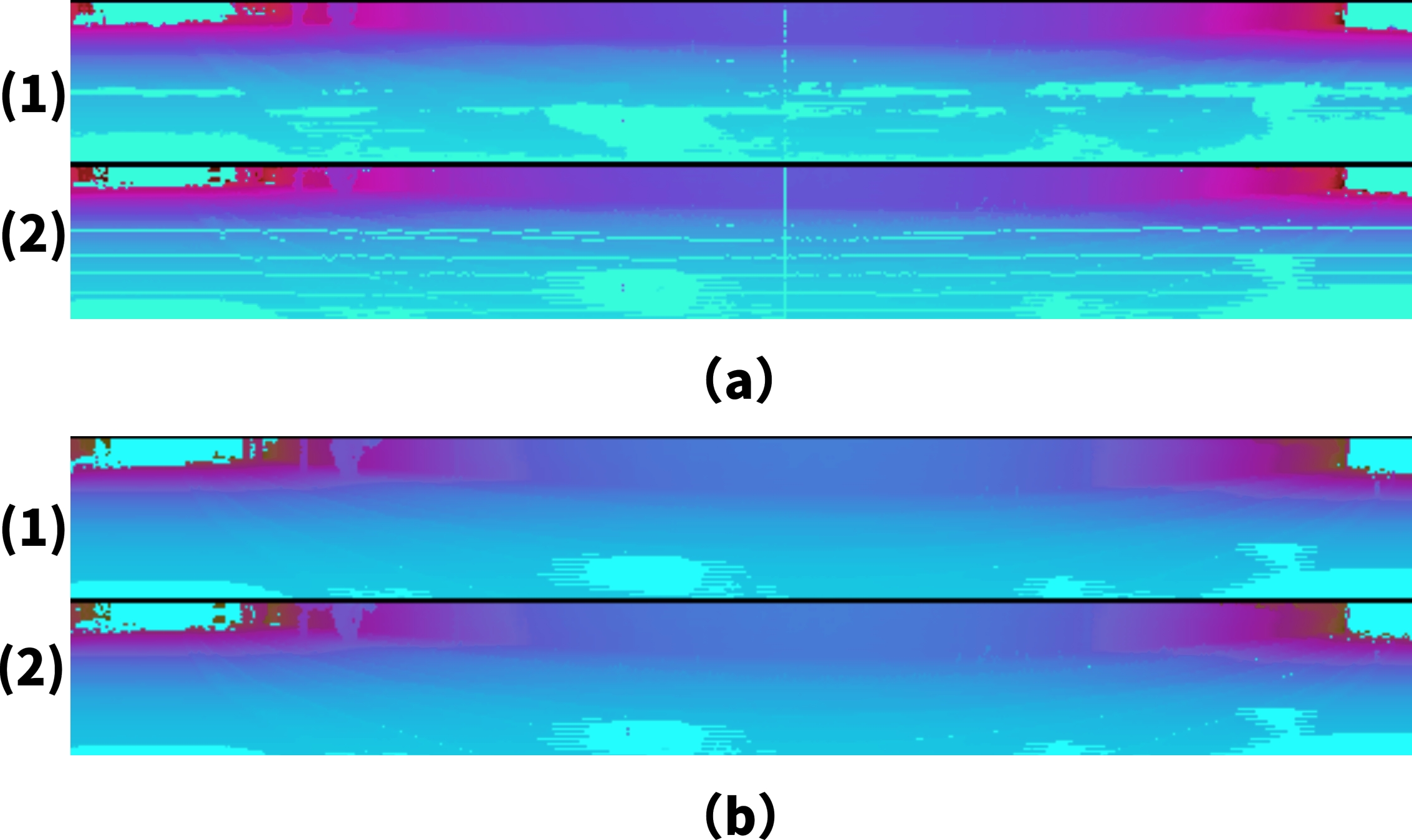}
\caption{(a) shows the original range image projection; (b) shows the improved range image quality after applying hough voting and hole compensation. (1)  denotes the corresponding projection  and (2) denotes the ground truth.}
\label{range-image-ablation}
\end{figure}

\begin{figure}[htbp]
\centering\includegraphics[width=\linewidth]{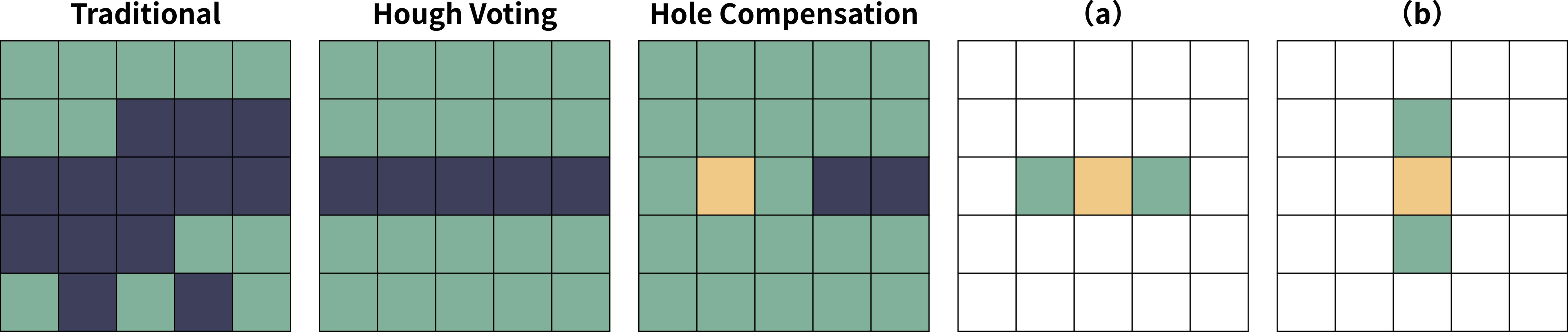}
\caption{Hough Voting. Green indicates valid pixels, black indicates hole regions, and yellow indicates pixel pooling regions. (a) and (b) respectively illustrate the pooling operations along the horizontal and vertical directions in the hole compensation mechanism. }
\label{hole-compensation}
\end{figure}

Large hole areas are inherent in the output of traditional range view methods. This characteristic imposes a fundamental limitation on the subsequent processing and makes image super-resolution techniques ineffective, as shown in Fig.~\ref{range-image-ablation}. The hough voting helps reduce structural breakage caused by hole pixels and maintains scene coherence. To further enhance the quality of the range view, a hole compensation mechanism is introduced, as illustrated in Fig.~\ref{hole-compensation}.

Table.~\ref{Hough-Voting and Hole Compensation} illustrates the effect of using different window shapes on the model performance. We take into consideration the pooling interpolation in horizontal and in vertical directions, respectively. The results show with vertical windowing strategy obtains better performance performance comparing to horizontal windowing. Point clouds are less affected by truncation errors in the horizontal direction, making horizontal pooling more prone to introduce additional noise, whereas in the vertical direction, point clouds exhibit similar feature distributions, and semantic information is smoother.

\begin{table}[htbp]
\centering
\caption{Ablation study on hough voting and hole compensation. The best-performing results are highlight in bold.}
\begin{tabular}{ccccc}
\hline
Methods & Hough Voting & Hole Compensation& IoU $\uparrow$  & CD $\downarrow$  \\
\hline
TULIP & \ding{55} & \ding{55} & 0.4152 & 0.1241 \\
TULIP & \ding{51} & \ding{55} & 0.4255 & 0.1068 \\
SRMamba-T & \ding{55} & \ding{55} & 0.4218 & 0.1198 \\
SRMamba-T & \ding{51} & \ding{55} & 0.4369 & 0.1068 \\
SRMamba-T & \ding{51} & \ding{51}($1\times 3$) & 0.4353 & 0.1080 \\
SRMamba-T & \ding{51} & \ding{51}($3\times 1$) & $\mathbf{0.4389}$ & $\mathbf{0.1031}$ \\
\hline
\end{tabular}
\label{Hough-Voting and Hole Compensation}
\end{table}

\subsubsection{Network Depth}

To verify the effect of network depth on the performance of SRMamba, we designs a set of ablation experiments on different depth configurations. Table.~\ref{depth} shows the quantitative evaluation results of SRMamba with different model depths, which further validates the effectiveness of the proposal method for multi-layer feature extraction.

\begin{table}[htbp]
\centering
\caption{Performance of neural networks with varying depths on point cloud super-resolution. }
\begin{tabular}{ccccc}
\hline
Depths & Params & CD$\downarrow$ & IoU $\uparrow$  & MAE $\downarrow$  \\
\hline
\texttt{SRMamba-T [2,2,2,2]} & 157.39M & 0.1031 & 0.4389 & 0.0044 \\
\texttt{SRMamba-S [2,2,9,2]} & 201.83M & 0.1018 & 0.4390 & 0.0044 \\
\texttt{SRMamba-M [2,2,12,2]} & 220.87M & 0.0982 & 0.4398 & 0.0055 \\
\texttt{SRMamba-L [2,2,27,2]} & 316.10M & 0.0940 & 0.4548 & 0.0048 \\
\hline
\end{tabular}
\label{depth}
\end{table}

\subsection{Failure Case}

\begin{figure}[htbp]
\centering\includegraphics[width=\linewidth]{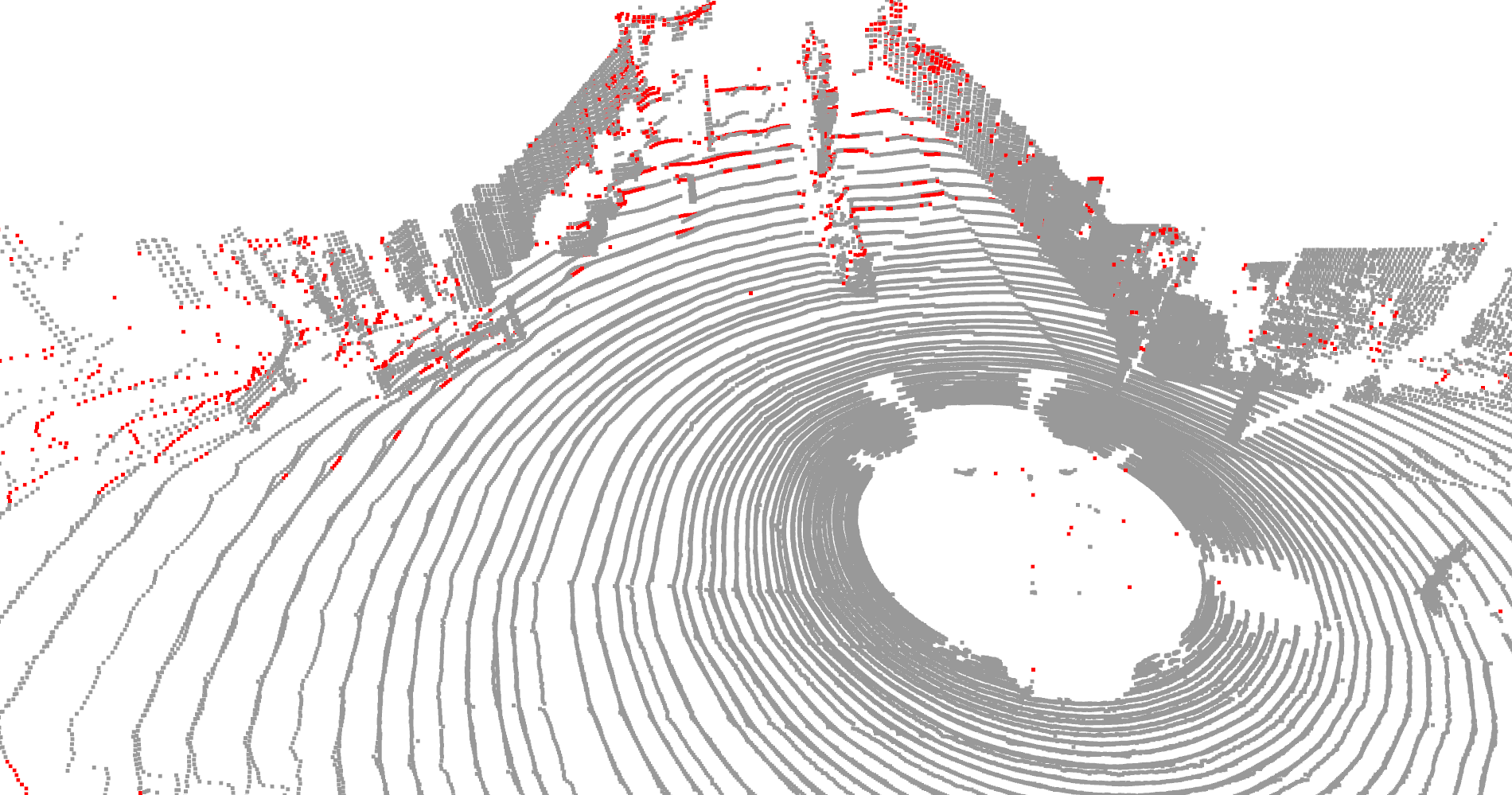}
\caption{Visualization image of the point cloud alignment results. Ground gray indicates areas matching the real point cloud, and red indicates areas with alignment errors exceeding 0.2 m.}
\label{Failure Case}
\end{figure}

Although the qualitative and quantitative evaluation results of SRMamba on the nuScenes \cite{fong2022panoptic} dataset are significantly better than those of other methods, the up-sampling of point clouds in sparse scenarios still faces a serious challenge, the problem also exists in the high-density KITTI-360 \cite{liao2022kitti} dataset. As shown in Fig.~\ref{Failure Case}, in the sparse edge region, our method still has some up-sampling errors. As the density of the point cloud decreases, the uncertainty of the spatial structure increases, leading to a further widening of the deviation between the reconstruction point cloud and the real point cloud.


\section{Conclusion}

This paper proposes a novel method, SRMamba, for large-scale low-resolution LiDAR point cloud super-resolution. The goal is to reconstruct realistic 3D scenes with lower computational cost. Unlike traditional approaches that rely on attention mechanisms for feature extraction, SRMamba uses a bidirectional scanning strategy based on sequence modeling to effectively capture long-range dependencies. It improves reconstruction quality under non-projection views while maintaining linear time complexity. Experiments on the KITTI and nuScenes datasets demonstrate its strong performance in both reconstruction accuracy and global modeling capability.

Future work will explore point cloud super-resolution in extremely sparse and challenging environments, such as rain, fog, and snow. The aim is to address environmental disturbances and enhance model robustness by improving generalization in complex real-world scenarios.

\begin{backmatter}
\bmsection{Funding}
was funded by the Key R\&D Project of the Sichuan Provincial Department of Science and Technology---Research on Three-Dimensional Multi-Resolution Intelligent Map Construction Technology (2024YFG0009), the Intelligent Identification and Assessment for Disaster Scenes: Key Technology Research and Application Demonstration (2025YFN0008), Project of the sichuan Provincial Department ofscience and Technology—— Application and Demonstration of Intelligent Fusion Processing of Laser Imaging Radar Data (2024ZHCG0176), the "Juyuan Xingchuan" Project of Central Universities and Research Institutes in Sichuan——High-Resolution Multi-Wavelength Lidar System and Large-Scale Industry Application(2024ZHCG0190), the Sichuan Science and Technology Program, Research on Simulator Three-Dimensional View Modeling Technology and Database Matching and Upgrading Methods, and~the Key Laboratory of Civil Aviation Flight Technology and Flight Safety (FZ2022KF08).

\end{backmatter}


\bibliography{sample}

\end{document}